\documentclass[a4paper,10pt,conference]{ieeeconf}
\IEEEoverridecommandlockouts
\usepackage{epsfig}
\usepackage{amsmath}
\usepackage{array}
\usepackage{amssymb}
\usepackage{cite}
\usepackage{booktabs}
\usepackage{multirow}
\usepackage{subcaption}
\usepackage{booktabs}
\usepackage{mathtools}
\usepackage{siunitx}
\usepackage{comment}
\usepackage{threeparttable}
\usepackage{bm}
\usepackage{pifont}
\makeatletter
\newcommand{\subsubsubsection}{\@startsection{paragraph}{4}{\z@}%
  {1.0\Cvs \@plus.5\Cdp \@minus.2\Cdp}%
  {.1\Cvs \@plus.3\Cdp}%
  {\reset@font\sffamily\normalsize}
}
\makeatother
\newcommand{\figlab}[1]{\label{fig:#1}}
\newcommand{\figref}[1]{Fig.~\ref{fig:#1}} 

\usepackage{color}
\usepackage{soul}

\usepackage{amsmath}
\usepackage{algorithmicx}
\usepackage[ruled]{algorithm}
\usepackage[noend]{algpseudocode}

\usepackage{xcolor}

\newif\ifshowrevisions
\showrevisionsfalse   

\ifshowrevisions
  \newcommand{\rev}[2]{{\color{blue}#2}\textsuperscript{\color{blue}\tiny[#1]}}
  \newenvironment{revblock}[1]
    {\par\noindent\textsuperscript{\color{blue}\tiny[#1]}\color{blue}}
    {\par}
\else
  \newcommand{\rev}[2]{#2}
  \newenvironment{revblock}[1]{}{}
\fi

\begin{document}
\title{Designing and Validating a Self-Aligning Tool Changer \\for Modular Reconfigurable Manipulation Robots}

\author{Mahfudz Maskur$^{1}$, Takuya Kiyokawa$^{1}$, and Kensuke Harada$^{1,2}$%
\thanks{$^{1}$Department of Systems Innovation, Graduate School of Engineering Science, The University of Osaka, 1-3 Machikaneyama, Toyonaka, Osaka, Japan.}%
\thanks{$^{2}$Industrial Cyber-physical Systems Research Center, The National Institute of Advanced Industrial Science and Technology (AIST), 2-3-26 Aomi, Koto-ku, Tokyo, Japan.}%
}

\maketitle

\begin{abstract}
Modular reconfigurable robots require reliable mechanisms for automated \rev{R2-1}{tool exchange}, but conventional rigid active couplings often fail due to inevitable positioning and orientational errors. To address this, we propose a misalignment-tolerant \rev{R2-1}{tool changer}. \rev{R2-2}{The tool changer consists of an insert part on the robot side, which contains a servo-driven cam and lock-pin, and a receptacle part on the tool side, which has an isosceles triangular lead-in guide and chamfered walls. During insertion, these geometries generate angled contact forces that passively align the two parts, so the tool changer compensates for angular and lateral misalignments without force sensors. A motor-driven rotating tool exchange station stores the tools and rotates to bring the selected tool to the exchange position. Experiments confirmed that, with the current guide and chamfer geometry, the tool changer tolerates rotational offsets of $\pm$20$^{\circ}$ and lateral offsets of up to 7\,mm.}
\end{abstract}

\IEEEpeerreviewmaketitle

\section{Introduction}
\rev{R2-3}{Modular reconfigurable robots consist of detachable modules and can change their morphology to suit various tasks and environments~\cite{Yim2007,Murata2002,Yim2000}. This paper focuses on their use in object manipulation.} A critical challenge in these systems is the development of reliable coupling mechanisms that enable \rev{R2-4}{automated tool exchange}. While non-actuated coupling mechanisms offer mechanical simplicity and natural compliance, they typically require highly precise external forces or manual intervention, thereby limiting full autonomy. Actuated couplings enable fully autonomous operation; however, they demand highly precise control to overcome positioning and orientational errors, presenting a significant challenge.

This becomes even more difficult because these systems typically operate with an unfixed-base. During autonomous manipulation and reconfiguration, small angular and lateral misalignments are practically inevitable due to kinematic inaccuracies, the inherent movements of the unfixed-base, or inertial effects during movement. Because conventional active couplings are typically rigid, they lack the passive compliance required to absorb these minor deviations. Consequently, they frequently suffer from engagement failures or mechanical jamming unless complex and expensive force/torque sensors are employed.

To overcome these limitations, this paper presents an automated, misalignment-tolerant \rev{R2-4}{tool changer}. Unlike conventional rigid couplings, our proposed \rev{R2-4}{motor-driven tool changer} incorporates passive self-aligning features, specifically chamfered receptacle walls and triangular lead-in guides. These features are designed to physically absorb both angular and lateral errors during the coupling process. Furthermore, to make this autonomous exchange practically feasible, the mechanism is complemented by a compact rotating tool exchange station for efficient \rev{R2-4}{tool} storage.

\begin{figure}[tb]
    \centering
    \includegraphics[width=\linewidth]{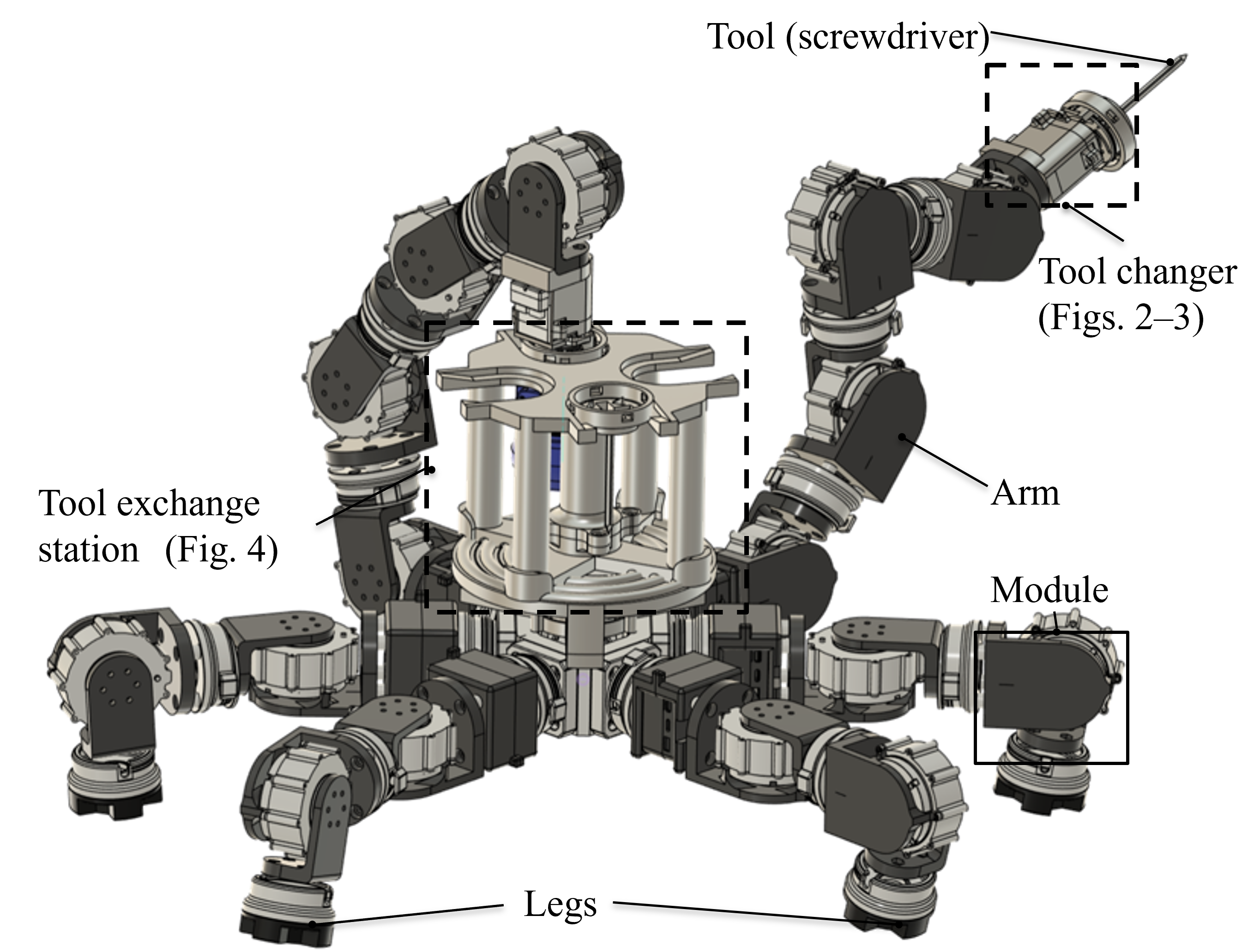}
    \caption{\protect\rev{R2-5}{CAD representation of the modular reconfigurable robot in a spider-shaped morphology. The robot is built from modules, each with a single function (e.g., a motor or a structural part); the limbs can serve as legs or arms depending on the configuration. The robot is designed to walk on its legs and can also be fitted with wheels, so no part is fixed to the environment. The arm carries a tool via the tool changer, and the tool exchange station on the robot's back stores the tools. Dashed boxes mark the parts detailed in Figs.~2--4.}}
    \figlab{spider}
\end{figure}

\begin{revblock}{R2-1}
In this paper, a \emph{module} is a detachable unit of the robot with a single function. A \emph{tool} is the end-effector at the tip of the robot; it can also be regarded as a module. The \emph{tool changer} connects the last module of the robot to a tool. It consists of the \emph{insert part}, mounted on the last module, and the \emph{receptacle part}, on the tool side. The \emph{tool exchange station} is a rotating table that stores the tools. We refer to detaching and attaching tools as \emph{tool exchange}.
\end{revblock}

\begin{revblock}{R2-10}
The intended application is assembly and disassembly lines in small and medium-sized factories. Large factories can install several robots, each dedicated to one task, whereas smaller factories may not be able to afford this. Instead, a single robot can move between workstations and perform several tasks by exchanging its tools, for example a gripper for handling parts and a screwdriver for fastening. Because the robot moves by itself, its base is not fixed, and the tool exchange station is carried on the robot, as shown in \figref{spider}. Locomotion itself is outside the scope of this paper.
\end{revblock}

The core contributions of this study are twofold: (1) the development of a misalignment-tolerant \rev{R2-4}{motor-driven tool changer} that utilizes passive self-aligning features to absorb physical positioning errors, and (2) the system-level integration of this mechanism with a compact rotating tool exchange station \rev{R3-2}{that is mounted on the robot and travels with it, so that the robot does not need to return to a fixed station, and that rotates to bring the selected tool within reach of the arm,} to demonstrate \rev{R3-2}{autonomous tool exchange in real-world experiments using standard MoveIt planning,} without the need for complex control or specialized algorithms.

\section{Related Work}

\subsection{\protect\rev{R2-4}{Robotic Couplings and Tool Changers}}
\rev{R2-4}{Tool changers} are widely adopted to expand a robot's operational capability across diverse tasks~\cite{Brown1999,Schlette2020}. Existing approaches are generally classified as passive or active. Passive systems~\cite{Pettinger2019,Berenstein2018} offer simplicity and compliance but typically require highly precise control strategies or external forces to engage and disengage \rev{R2-4}{tools}~\cite{Yi2023}. Conversely, active methods enable greater autonomy through built-in actuation~\cite{Li2022,Makabe2024,Roux2024}. However, conventional active designs often rely on rigid interfaces. When subjected to the inevitable positioning errors of autonomous operation, these rigid couplings are prone to failure unless augmented with complex compliance controllers or force/torque sensors. 

In contrast to these conventional approaches, our work introduces a hybrid concept: an active, servo-driven \rev{R2-4}{tool changer} embedded with passive self-alignment geometry. By utilizing chamfered walls and triangular lead-in guides, our system achieves the autonomy of active mechanisms while providing the misalignment tolerance typically only found in specialized compliant passive systems, \rev{R3-1}{without force sensors}.

\subsection{Reconfiguration Planning in Modular Systems}
Self-reconfigurable robots have evolved significantly from basic locomotion and self-assembly~\cite{Murata2002,Yim2000,Devey2012, Meibao2019,Gerbl2022,Bhattacharjee2022,Zhao2025} to highly mobile and adaptable robots~\cite{Liu2023,Sprowitz2010}. However, transitioning these systems toward practical object manipulation requires mechanisms that offer high joint stiffness while retaining reliable detachability. 

Furthermore, while trajectory planning and motion synthesis have been extensively studied for general manipulation~\cite{Leslie2013} and dynamic locomotion~\cite{Carlo2018}, planning specifically for the physical coupling phase in modular robots remains challenging. Prior works have proposed heuristics for structural stability~\cite{Yim2007} and task-driven configuration synthesis~\cite{Icer2016}. Yet, ensuring reliable coupling during the actual runtime tool-exchange process, where planner approximations encounter real-world physical discrepancies, remains underexplored.
        
\rev{R3-2}{For example, in~\cite{Berenstein2018}, the robot is mobile, but the tool station is fixed. In our system, the rotating tool exchange station is mounted on the robot and travels with it, so the robot does not need to return to a fixed station, and the station rotates to bring the selected tool within reach of the arm. Trajectories are planned with standard MoveIt, without a specialized planner for the coupling phase; the positional deviations that arise during execution are absorbed physically by the self-aligning tool changer (Sec.~IV-D).}

\section{\protect\rev{R2-9}{Proposed Mechanism}}

This section presents the design of \rev{R2-4}{the motor-driven tool changer} and the rotating tool exchange station.

\subsection{\protect\rev{R2-4}{Tool Changer}}

\begin{figure}[tb]
    \centering
    \includegraphics[width=\linewidth]{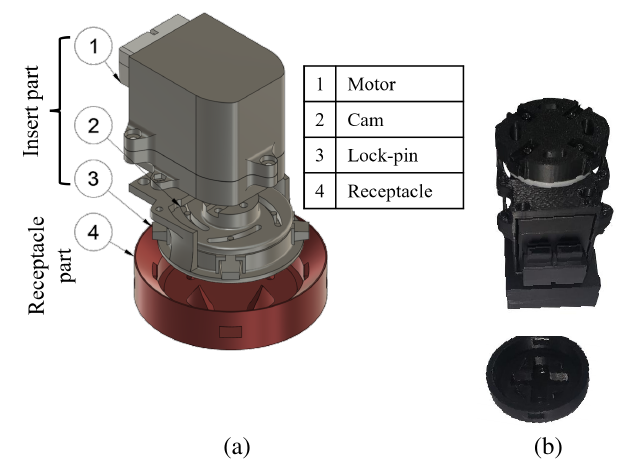}
    \caption{\protect\rev{R2-6}{Motor-driven tool changer: (a) CAD representation, consisting of the insert part (motor, cam, and lock-pin) and the receptacle part; (b) 3D-printed insert part (top), shown with the connector to the last module of the arm, and receptacle part (bottom). It corresponds to the dashed box at the arm tip in Fig.~1.}}
    \figlab{coupling-explode}
\end{figure}

\begin{figure}[tb]
    \centering
    \includegraphics[width=\linewidth]{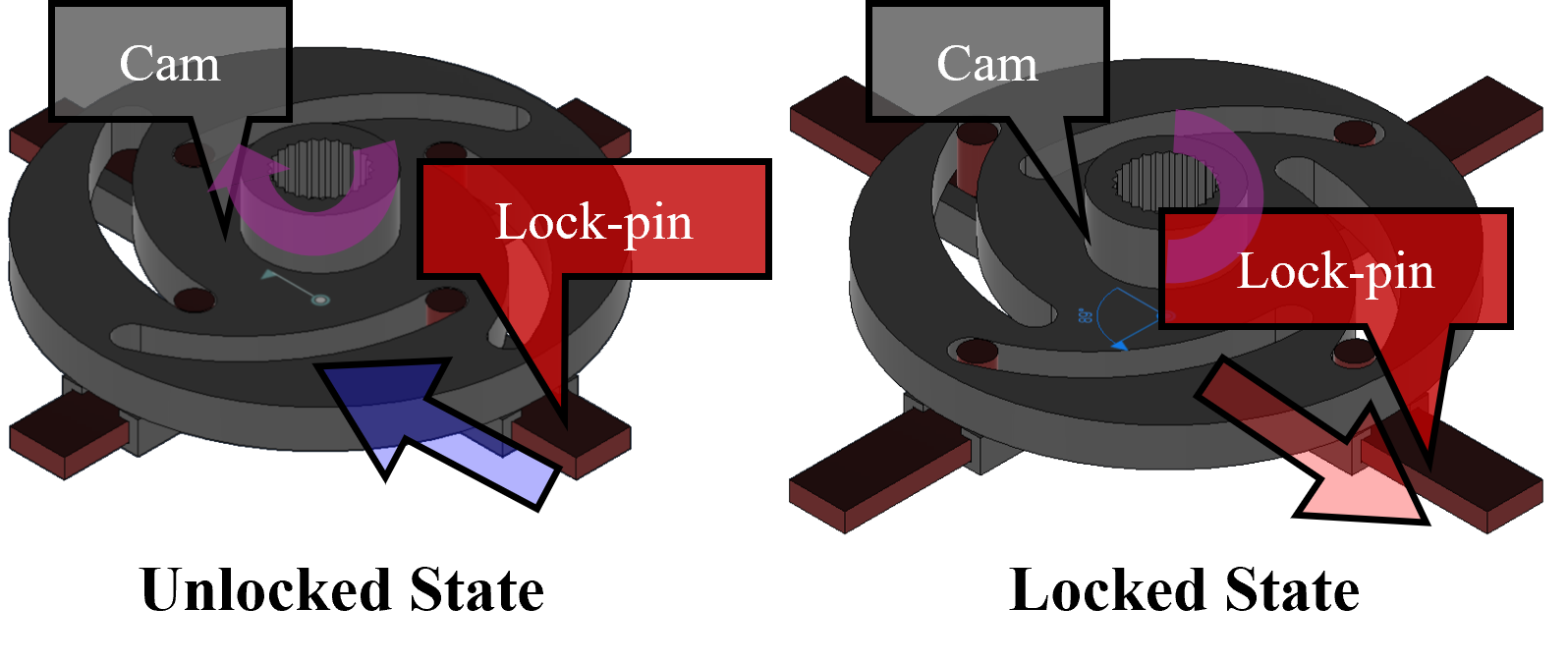}
    \caption{\protect\rev{R2-1}{Cam-lock mechanism of the tool changer. In the unlocked state, the cam pulls the lock-pin inward; in the locked state, the cam rotates in the opposite direction and pushes the lock-pin outward. The blue lines indicate the cam rotation.}}
    \figlab{coupling-mechanism}
\end{figure}

\figref{coupling-explode}\rev{R2-6}{(a)} shows \rev{R2-1}{a CAD view of the motor-driven tool changer with its main components: the servo motor (KRS-9304HV ICS), the cam, and the lock-pin in the insert part, and the receptacle part.} \rev{R2-6}{\figref{coupling-explode}(b) shows the 3D-printed parts. The photo of the insert part also includes the connector to the last module of the arm, which is omitted in the CAD view.}
The insert contains the servo motor, cam mechanism, and lock-pin inside a mounting case attached to the \rev{R2-4}{last module of the robot}. This part includes an opening for the lock-pin when mating with the \rev{R2-4}{receptacle part}.

The receptacle is designed to accept the lock-pin and features a circular triangle lead-in guide. This protruding geometry ensures correct orientation, guiding the lock-pin for reliable engagement.

\figref{coupling-mechanism} shows the \rev{R2-4}{tool changer} using a motor-driven cam-lock system as its core. When actuated, the servo rotates the cam shaft between 0 degrees (unlocked) and 60 degrees (locked), converting rotary motion into linear movement of the lock-pin. The lock-pin, attached with a screw linkage, extends to latch onto the \rev{R2-4}{receptacle part} and secure the connection.
The receptacle's basic hole accommodates both the lock-pin and the lead-in guide. This design can be modified with passive attachment features for different \rev{R2-4}{tools}. While explicit locking tabs are not used, the cam-lock pin system provides strong mechanical retention and prevents accidental decoupling.

\subsection{Rotating Tool Exchange Station}
\begin{figure}[tb]
    \centering
    \includegraphics[width=\linewidth]{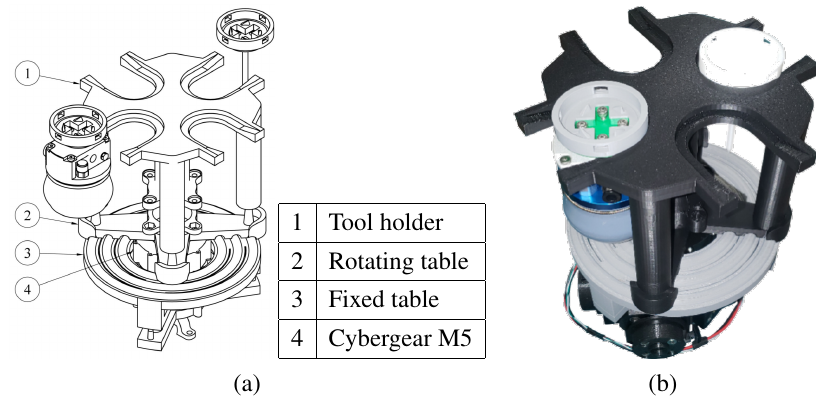}
    \caption{\protect\rev{R2-6}{Rotating tool exchange station: (a) CAD representation and its main components; (b) 3D-printed station. It corresponds to the dashed box on the robot's back in Fig.~1.}}
    \figlab{table-explode}
\end{figure}

\rev{R2-6}{\figref{table-explode}(a) shows a CAD view of the station with its main components, and \figref{table-explode}(b) shows the 3D-printed station.}
The station is composed of a rotating table that holds \rev{R2-4}{the tools}. Instead of using clamps or locking mechanisms, the design incorporates simple geometric features, including small triangular extrusions on the table surface. These extrusions help guide and position the \rev{R2-4}{tools} accurately during storage and retrieval.

Rotation is provided by a Cybergear M5 motor, chosen for its higher torque compared to the servo used in the \rev{R2-4}{tool changer}. This added torque allows the table to reliably rotate, even when handling heavier \rev{R2-4}{tools}. The robot can then access the desired \rev{R2-4}{tool} by rotating the table to the appropriate position.

\section{Experimental Evaluation}
This section presents an experimental evaluation of the proposed \rev{R2-4}{tool changer and tool exchange station}. The evaluation focuses on two aspects: tolerance to misalignment during \rev{R2-4}{insertion} and validation of \rev{R2-4}{tool exchange} using a modular robot.

\subsection{\protect\rev{R1-1}{Misalignment Problem}}
\begin{figure}[tb]
    \centering
    \includegraphics[width=\linewidth]{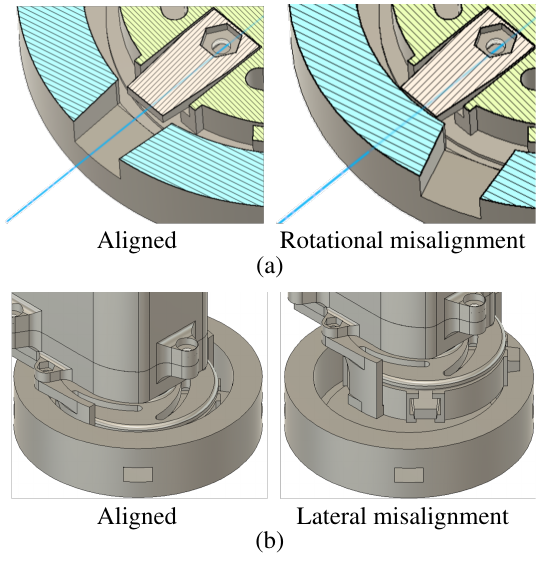}
    \caption{\protect\rev{R1-1}{Misalignment between the insert part and the receptacle part: (a) rotational and (b) lateral. Left: aligned; right: misaligned.}}
    \figlab{rot-mis}
\end{figure}

\figref{rot-mis}\rev{R1-1}{(a)} illustrates the rotational alignment challenges encountered during insertion when the lock-pin cannot be engaged. \figref{rot-mis}\rev{R1-1}{(b)} illustrates the lateral alignment challenges encountered during insertion when the \rev{R2-4}{insert part} cannot be successfully engaged due to positional offset.

To maintain consistency with the physical constraints of the robot's modular links, the receptacle wall thickness was fixed at \rev{R3-3}{7.5\,mm} in all alignment experiments.
\rev{R3-3}{The tested receptacle has an outer diameter of 65\,mm. The inner surface of its wall has a 45$^{\circ}$ chamfer that is 7\,mm wide. Four triangular lead-in guides stand on the receptacle floor, 90$^{\circ}$ apart, and each spans 40$^{\circ}$ of arc. The isosceles guide is centred on this span, whereas the right-angled guide is aligned with one side of it.} \rev{R1-5}{\figref{receptacle-geometry} shows the guides and the chamfered wall in a cross-section.}

\begin{figure}[tb]
    \centering
    \includegraphics[width=\linewidth]{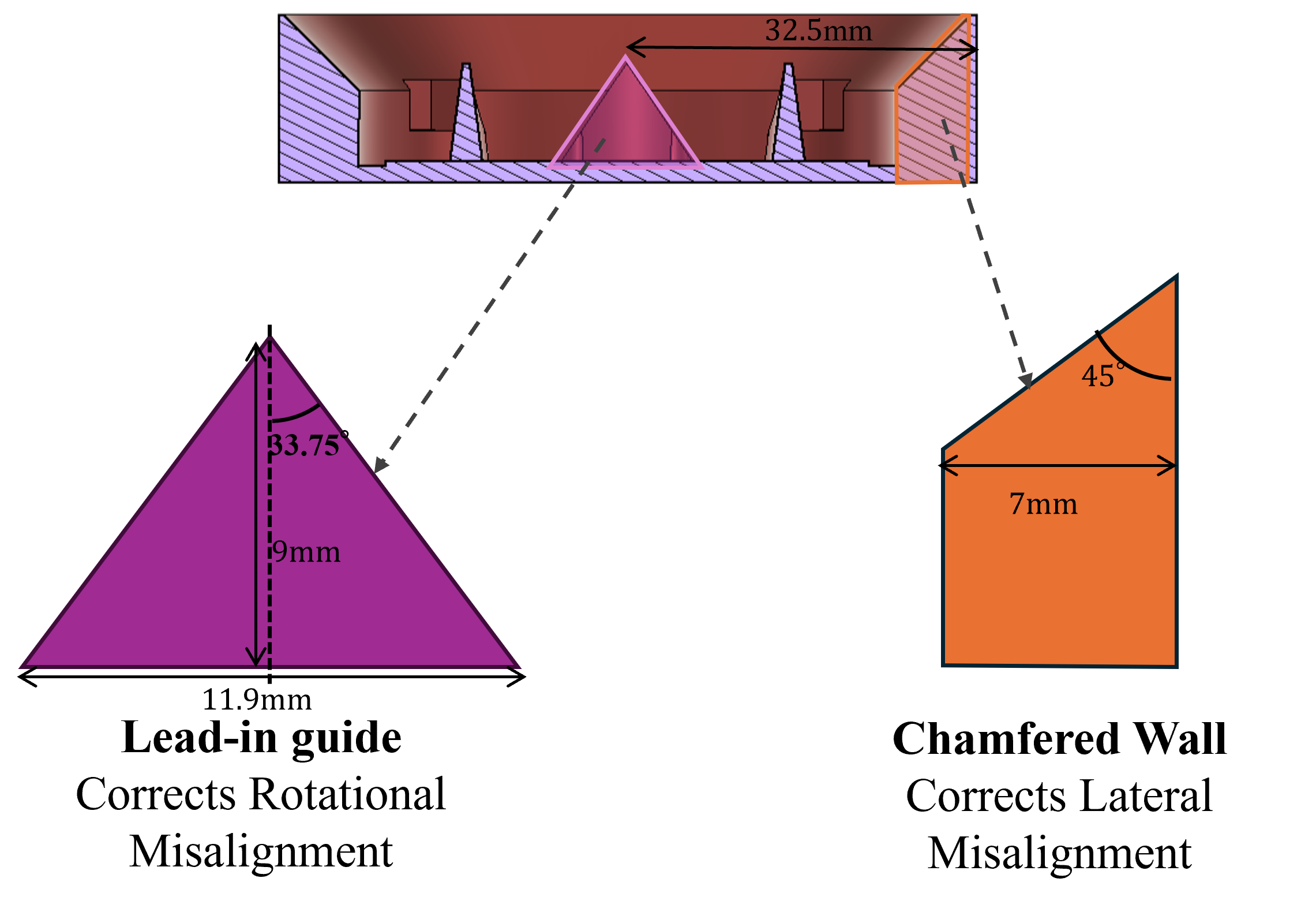}
    \caption{\protect\rev{R1-5}{Cross-section of the receptacle part: the lead-in guides on the floor correct rotational misalignment, and the chamfered wall corrects lateral misalignment.}}
    \figlab{receptacle-geometry}
\end{figure}

\begin{figure}[tb]
    \centering
    \includegraphics[width=\linewidth]{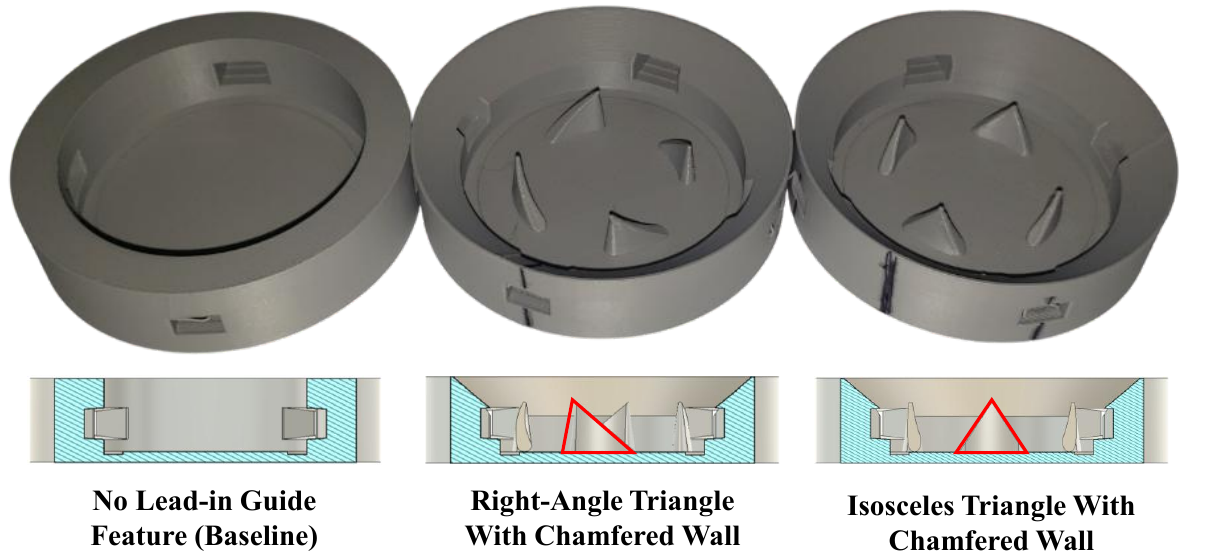}
    \caption{CAD representation and the 3D printed of the examined receptacle.}
    \figlab{exam-plate}
\end{figure}

\figref{exam-plate} shows the CAD design and 3D-printed receptacles used in these experiments. For the angular tests, two lead-in guide geometries were tested: right-angled and isosceles triangles, each with chamfered walls. Additionally, a baseline receptacle without any lead-in guide was evaluated for comparison. For the lateral tests, the receptacle employed the isosceles triangle lead-in guide, comparing chamfered and non-chamfered wall designs.

\subsection{\protect\rev{R1-1}{Misalignment Tolerance by Manual Insertion}}
Experiments were performed by manual insertion with five trials per condition. The offset was gradually increased from an initially aligned configuration to determine the maximum tolerance.

\begin{revblock}{R1-1}
In the manual tests, the insert part was held by hand and pushed straight down, while the receptacle was placed freely on a reference plate so that it could slide and rotate. The starting position and orientation of the receptacle were marked on the plate. Angular offsets were set by rotating the receptacle according to printed angle marks on the plate, and lateral offsets by sliding it according to printed marks at 1\,mm intervals, in both directions in each case. A trial was counted as successful when the insert part was fully seated and the lock-pin entered the receptacle; the receptacle then returned to the starting mark. A trial failed when the insert part was blocked by the receptacle wall or the lead-in guide and could not enter the receptacle.
\end{revblock}

\subsubsection{Angular misalignment}
The offset angle was incremented by 2.5$^{\circ}$ in both directions until insertion failure.

\begin{figure}[tb]
    \centering
    \includegraphics[width=\linewidth]{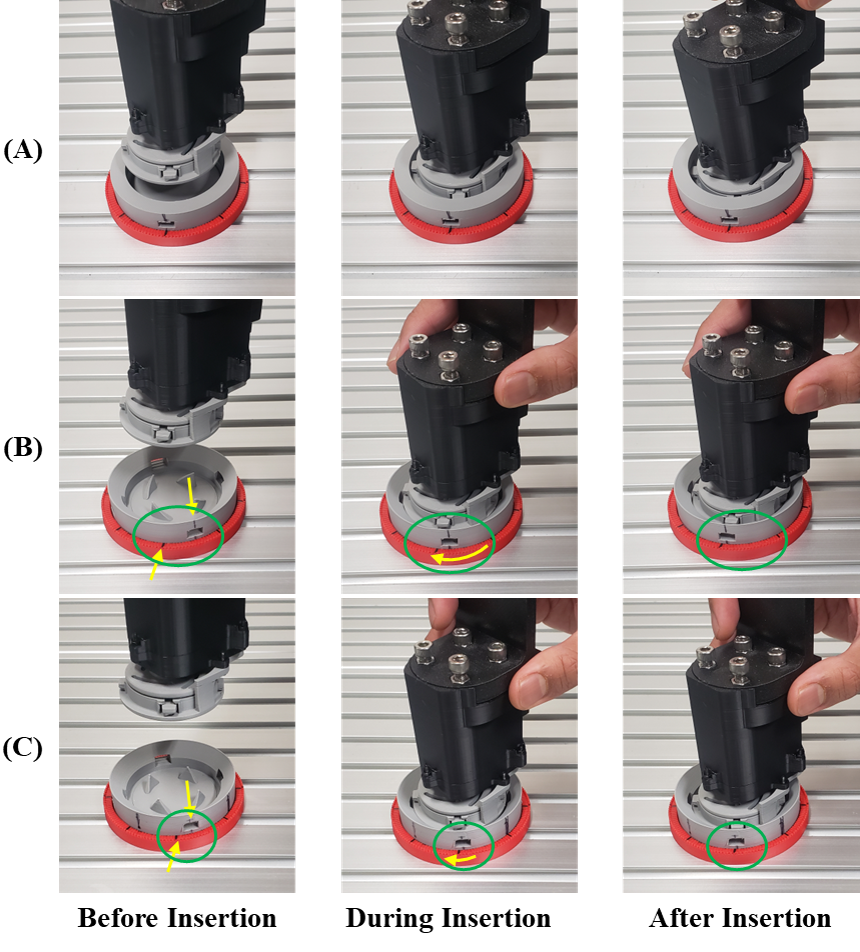}
    \caption{Step-by-step insertion of the motor-driven \protect\rev{R2-4}{tool changer} under three conditions: (A) no lead-in guide feature as a baseline, (B) right-angle triangle with chamfered wall, and (C) isosceles triangle with chamfered wall. Images from left to right show the states before, during, and after insertion.}
    \figlab{angular-exam}
\end{figure}

In \figref{angular-exam}, the receptacles with lead-in guides rotate into the correct orientation during insertion, as seen from the marks on the red reference plate.
The baseline receptacle without lead-in guide successfully coupled only at the correct alignment angle.
The right-angled triangle lead-in guide enabled coupling with angular misalignments from 0$^{\circ}$ to +40$^{\circ}$.
The isosceles triangle lead-in guide enabled coupling with angular misalignments \rev{R1-1}{from $-$20$^{\circ}$ to +20$^{\circ}$}.
\rev{R1-1}{All five trials succeeded at $\pm$20$^{\circ}$, and all five failed at $\pm$22.5$^{\circ}$.}
\rev{R3-3}{Both ranges equal the 40$^{\circ}$ span of the guide, which is 44\,\% of the 90$^{\circ}$ symmetry period.}
\subsubsection{Lateral misalignment}
The lateral offset was increased in increments of 1 mm in both lateral directions until insertion failure occurred.

\begin{figure}[tb]
    \centering
    \includegraphics[width=\linewidth]{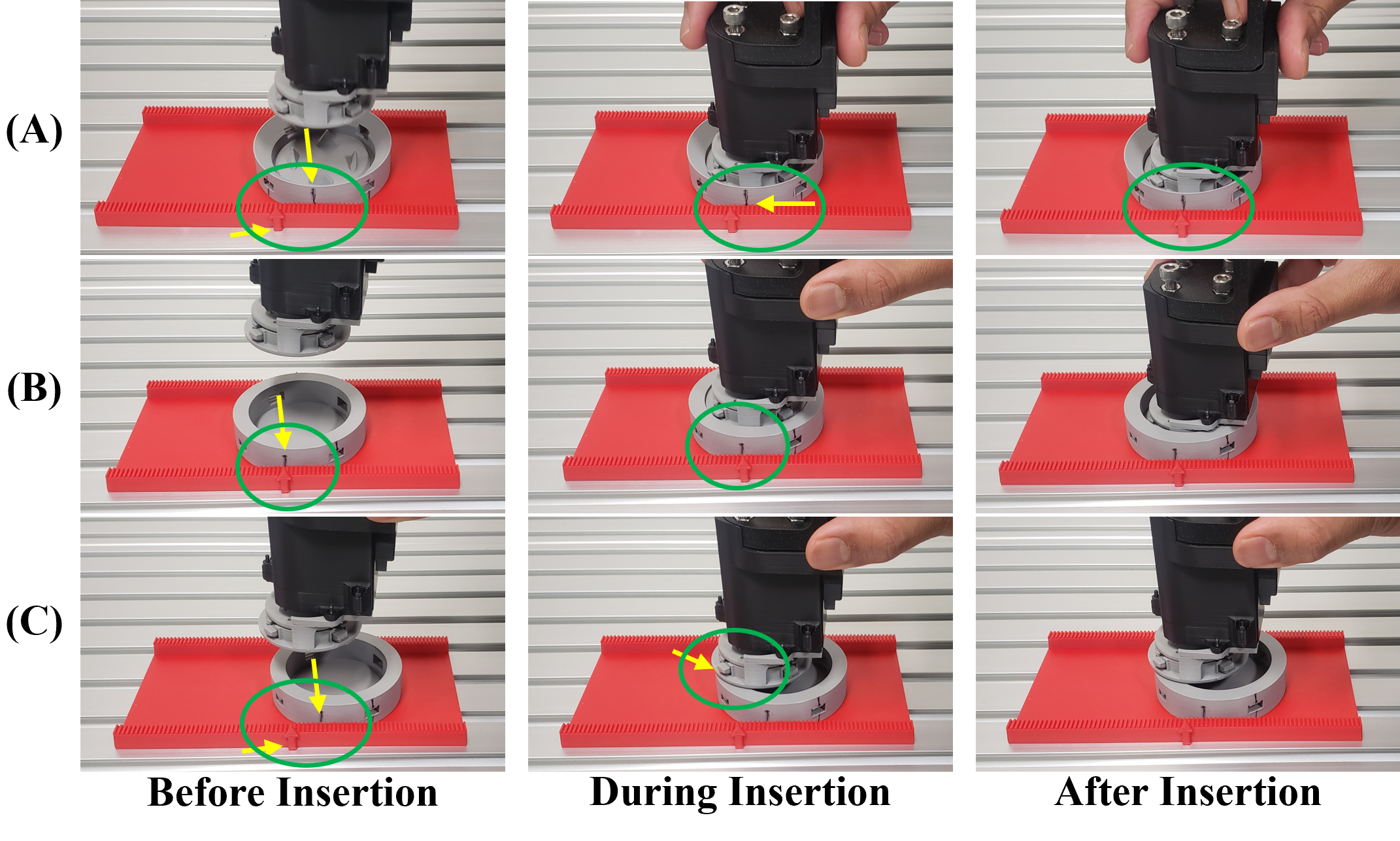}
    \caption{Insertion sequence under different receptacle conditions: (A) isosceles triangle with chamfered wall, (B) baseline with correct alignment, and (C) baseline with lateral misalignment. Images from left to right show the states before, during, and after insertion.}
    \figlab{lateral-exam}
\end{figure}

In \figref{lateral-exam}, the receptacle with chamfered walls slides into the correct position during insertion, as seen from the marks on the reference plate.
The baseline receptacle without a chamfered wall was only able to couple successfully at the correct lateral alignment position.
The chamfered wall design enabled coupling with lateral offsets of up to \rev{R1-1}{7\,mm}.
\rev{R1-1}{The same result was obtained in both directions: all five trials succeeded at 7\,mm, and all five failed at 8\,mm.}
\rev{R3-3}{This tolerance equals the 7\,mm chamfer width, which corresponds to 93\,\% of the 7.5\,mm wall thickness and 11\,\% of the 65\,mm outer diameter.}

\subsection{\protect\rev{R1-1}{Validation with an Industrial Robot}}
\begin{revblock}{R1-1}
To check the manual results with precisely set offsets, the tests with the isosceles guide and chamfered walls were repeated using a KUKA LBR iiwa robot. The insert part was mounted on the robot flange with a 3D-printed adapter plate and bolts, and the receptacle was placed loosely on the plate. The robot was position-controlled, and each offset was applied by adding it to the commanded pose of the robot end-effector; no force control was used. For safety only, the program stopped the robot when the contact force exceeded 500\,N, and such a stop was counted as a failure. The robot tests confirmed the manual results: insertion succeeded up to $\pm$20$^{\circ}$ and 7\,mm and failed beyond these offsets. Each step was tested three times (0$^{\circ}$ row and 0\,mm column in \figref{combined}).
\end{revblock}

\begin{revblock}{R3-4}
For a quantitative comparison of the alignment during insertion, a commercial manual tool changer (Kosmek SXR), whose receptacle has an outer diameter of 63\,mm, was tested with the same robot setup and procedure, with angular steps of 1$^{\circ}$. Because its lock is operated by hand, a trial was counted as successful when the tool adapter was fully seated. In the proposed tool changer, a fully seated receptacle can always be locked, so the same criterion applies to both. It aligns through a single cone-shaped pin, placed off the centre on one side, and guide features in the receptacle wall near the locking elements. It tolerated angular offsets of $\pm$9$^{\circ}$ and lateral offsets of $\pm$3\,mm, compared with $\pm$20$^{\circ}$ and $\pm$7\,mm for the proposed tool changer. Normalized by the outer diameter, the lateral tolerance is 5\,\% for the commercial tool changer and 11\,\% for the proposed one. In addition, because of the single pin, the commercial tool changer couples in only one orientation per 360$^{\circ}$, whereas the proposed tool changer repeats every 90$^{\circ}$. Three trials were performed at the steps near the tolerance limits.
\end{revblock}

\begin{revblock}{R1-2}
Both offsets were then applied at the same time (5--20$^{\circ}$ in 5$^{\circ}$ steps, 1--7\,mm in 1\,mm steps). Each condition was repeated until three trials succeeded, or stopped after three failures. As shown in \figref{combined}, the tolerated lateral offset decreases as the angular offset increases.
\end{revblock}

\begin{figure}[tb]
    \centering
    \includegraphics[width=\linewidth]{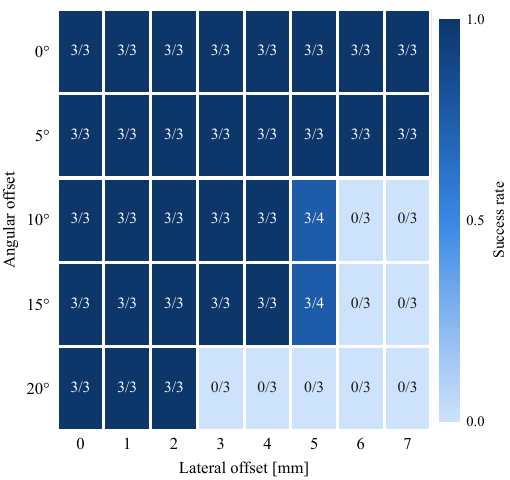}
    \caption{\protect\rev{R1-2}{Success rate for combined angular and lateral offsets (robot arm, isosceles guide with chamfered walls). Each cell shows successful trials over attempts; the 0$^{\circ}$ row and the 0\,mm column are single offsets.}}
    \figlab{combined}
\end{figure}

\subsection{\protect\rev{R2-4}{Tool Exchange Experiment}}
\rev{R2-4}{To evaluate system-level performance, a 6-axis articulated arm was constructed from the proposed modules, with the tool changer mounted on its last module, and a fully automated tool exchange sequence was conducted with the tool exchange station. Trajectories were planned in MoveIt to obtain collision-free paths and repeatable insertion conditions. As shown in \figref{simulation}, the planned trajectory guides the arm from a fully extended initial pose to the tool exchange position. The same trajectory was reused in the hardware experiments, although repeated trials were initialized from the disengaged pose for practicality.}

\begin{figure*}[tb]
    \centering
    \includegraphics[width=\linewidth]{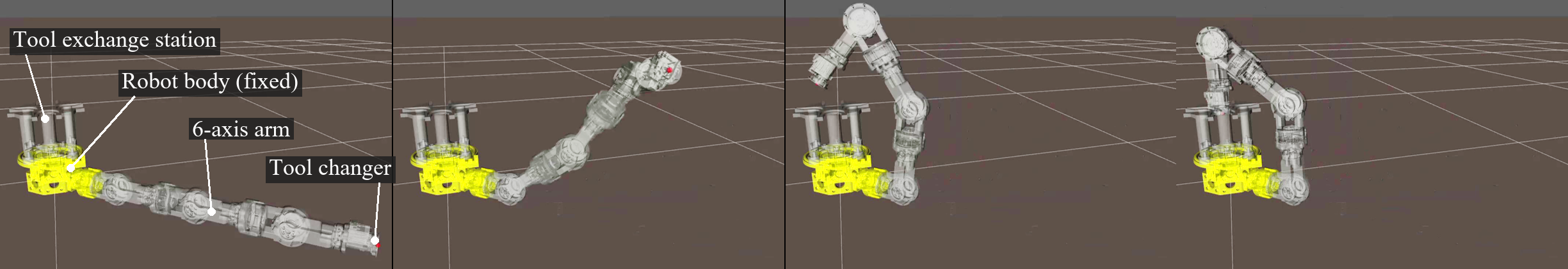}
    \caption{Time-lapse sequence of the simulation environment. \protect\rev{R2-7}{The robot body (yellow) is fixed, and the legs are omitted in the simulation.} The 6-axis modular arm executes a planned trajectory from a fully extended initial pose (left) to the tool exchange position (right).}
    \figlab{simulation}
\end{figure*}

The receptacle was mounted on a rotating table, and repeated tool attachment and detachment cycles were performed. \rev{R3-2}{The station was mounted on the back of the robot, which did not walk during the experiment, and held one receptacle. In MoveIt, rotating the station was necessary for the arm to reach some of the tool positions.} \figref{experiment} illustrates the tool exchange sequence. In each trial, the \rev{R2-4}{insert part} disengaged, moved upward, returned to the exchange position, and was then re-inserted to achieve passive self-alignment before locking and lifting the \rev{R2-4}{receptacle part}.

\begin{figure*}[tb]
    \centering
    \includegraphics[width=\linewidth]{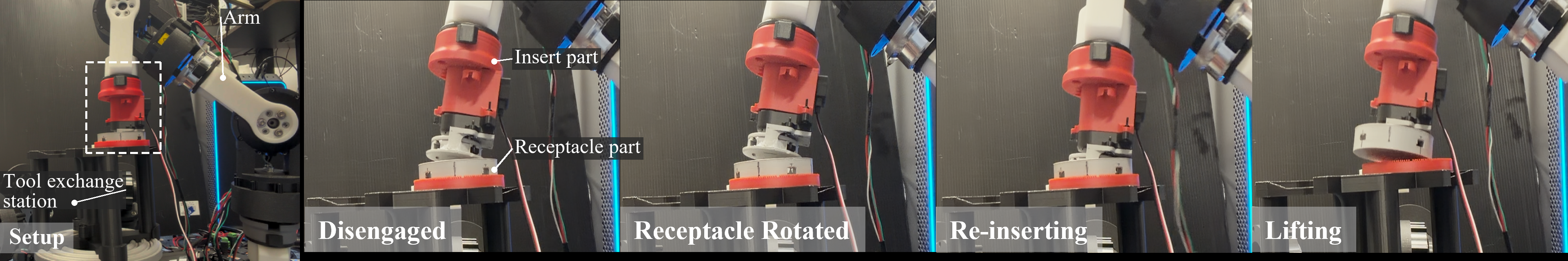}
    \caption{Autonomous tool exchange sequence. \protect\rev{R2-8}{Setup: the 6-axis modular arm holds the insert part above the tool exchange station; the robot was placed on a table. The dashed box marks the area shown in the close-ups.} Close-ups: disengagement, receptacle rotation to introduce misalignment, re-insertion with passive self-alignment, and lifting after successful locking. \protect\rev{R2-8}{The red plate under the receptacle is a reference plate for checking the misalignment.}}
    \figlab{experiment}
\end{figure*}

To evaluate the effect of passive self-alignment during autonomous operation, two conditions were tested. In the first condition, the receptacle remained fixed after disengagement. In the second condition, the receptacle was intentionally rotated \rev{R1-6}{by hand, by an estimated angle within the measured tolerance of $\pm$20$^{\circ}$,} to introduce \rev{R1-6}{orientational} misalignment prior to re-insertion. Each condition was tested for ten trials. \rev{R1-6}{In both conditions, an additional lateral and angular offset of unknown size was present because of the positioning error of the modular arm, whose modules are joined by detachable connectors (\figref{arm-offset}).}

As a result, the system achieved 10 successful tool exchanges out of 10 trials when the receptacle position was fixed, and 9 successful exchanges out of 10 trials when the receptacle was intentionally rotated.
\rev{R1-6}{A trial was counted as a failure when the insert part could not be fully seated in the receptacle; no failure occurred during locking or lifting. In the single failed trial, the insert part landed on the receptacle wall and was blocked. The most likely cause is that the intentional rotation and the positioning error of the arm together exceeded the tolerance of the tool changer.}

\begin{figure}[tb]
    \centering
    \includegraphics[width=\linewidth]{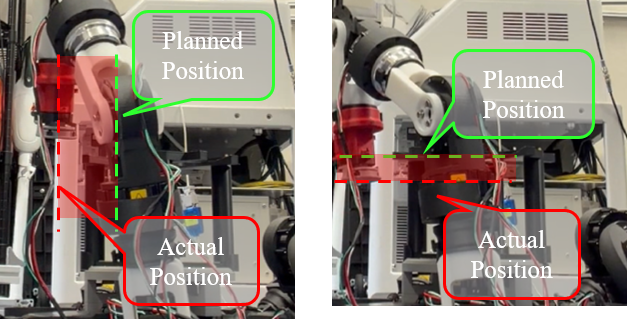}
    \caption{\protect\rev{R1-2}{Offset between the planned and the actual position of the modular arm during the tool exchange trials, shown in two views.}}
    \figlab{arm-offset}
\end{figure}

\begin{revblock}{R1-2}
As shown in \figref{arm-offset}, the actual position of the modular arm differed from the position planned in MoveIt. The arm sagged because of gravity and the limited joint torque of the modules. Increasing the control gain of the joint motors reduced this offset, but a visible offset remained. This remaining offset was absorbed by the tool changer during insertion. The offset was not measured quantitatively.
\end{revblock}

\section{\protect\rev{R1-4}{Discussion}}
\textbf{Comparison with prior tool changers.} \rev{R3-1}{Chamfers and lead-in guides are common in tool changers. In the tested commercial tool changer, the guiding features are a single alignment pin and local guides near the locks. In the proposed tool changer, the chamfer covers the whole circumference of the receptacle wall, and the guides are arranged around the rotation axis. At a similar size, this gives larger lateral and angular tolerances, as shown in Sec.~IV-C.}

\textbf{Uncovered region around 45$^{\circ}$.} \rev{R1-3}{The four guides cover 40$^{\circ}$ of every 90$^{\circ}$. If the rotational offset lies between 20$^{\circ}$ and 70$^{\circ}$, i.e., in a 50$^{\circ}$ region centred on 45$^{\circ}$, the insert part lands on top of a guide and is blocked. The current system cannot detect or correct such an offset; it relies on the tools remaining close to their storage orientation, since the triangular extrusions on the tool exchange station limit lateral sliding of the tools but not their rotation. Wider guides would reduce this region, but at the same receptacle size they leave less room for the lock-pin, or they require a thicker receptacle, whereas we aim for a size comparable to commercial tool changers such as the Kosmek SXR. Fewer guides, e.g., three, would allow a larger span per guide, but give fewer lock orientations and less stability; this trade-off has not been tested. The isosceles guide is preferred over the right-angled guide because it corrects in both directions, whereas the right-angled guide fails even at $-$2.5$^{\circ}$. Detecting the tool orientation with a sensor, or adding a module that rotates the insert part about its axis, would bring the offset into the covered range; this is left for future work.}

\textbf{Assumptions and limitations.} \rev{R1-4}{The tool changer assumes that the insert part approaches straight along the coupling axis; tilted insertion was not tested. After locking, a small play remains between the lock-pin and the receptacle, because the parts are 3D-printed from PLA with limited precision. A larger clearance makes locking easier, so this is a trade-off; the play was not measured, and a more precise lock-pin is expected to reduce it. Because the tool can lock in any of four orientations, 90$^{\circ}$ apart, and the robot cannot detect which one, the current design suits rotationally symmetric tools such as a screwdriver, a suction gripper, or a jamming gripper; asymmetric tools would require a sensor to detect the orientation. Failed insertions can deform the PLA lead-in guide. In the reported trials, no tool was mounted on the receptacle, and the load capacity was not tested. Compared with purely passive mechanisms, the tool changer requires a motor and power on the robot side, but it does not rely on external forces for engagement and release, and the cam remains locked when power is lost. Compared with pneumatic tool changers, it needs no air supply or airtight parts, which are difficult to fabricate by 3D printing. Finally, the positioning error of the modular arm was observed but not measured.}

\section{Conclusion}

This paper presented the experimental validation of a self-aligning \rev{R2-4}{tool changer} designed for autonomous tool exchange. By utilizing passive geometric features, specifically triangular lead-in guides and chamfered receptacle walls, the proposed interface \rev{R1-1}{compensates for angular offsets of $\pm$20$^{\circ}$ and lateral offsets of up to 7\,mm}. This design \rev{R3-1}{reduces the need for highly precise control and avoids force sensing or active compliance control}. \rev{R3-4}{In a comparison under the same conditions, a commercial manual tool changer tolerated $\pm$9$^{\circ}$ and $\pm$3\,mm.}

System-level experiments using a 6-axis modular arm confirmed \rev{R3-1}{this approach in practice}, achieving a 10/10 success rate with a fixed receptacle and a 9/10 success rate under intentionally introduced misalignment. Overall, these results demonstrate that the proposed mechanism \rev{R3-1}{enables} automated \rev{R2-4}{tool exchange} in unfixed-base modular robots where maintaining high precision is difficult. \rev{R1-4}{Limitations include the uncovered angular region around 45$^{\circ}$ and the unmeasured positioning error of the modular arm.} Future work will investigate long-term durability, \rev{R1-4}{add orientation sensing for asymmetric tools and the 45$^{\circ}$ region, measure the positioning error of the modular arm,} expand the range of interchangeable tools, and evaluate the system's performance not only in tool exchange but also in full morphological reconfiguration scenarios.

\section*{Acknowledgement}
This work was financially supported by the Nagamori Foundation.

\bibliographystyle{IEEEtran}
\bibliography{references}

\end{document}